\documentclass{article}
\usepackage[T1]{fontenc}
\usepackage[utf8]{inputenc}  
\usepackage{lmodern}
\usepackage{xcolor}
\usepackage[most]{tcolorbox}
\usepackage[preprint]{corl_2026} 
\usepackage{graphicx}
\usepackage{booktabs} 
\usepackage{amsmath} 
\usepackage{amssymb}
\usepackage{enumitem}
\usepackage{hyperref}
\usepackage{wrapfig}
\usepackage{multirow}

\usepackage{makecell}
\usepackage{pifont}

\definecolor{deepgreen}{RGB}{0,128,0}
\definecolor{deepred}{RGB}{160,0,0}

\definecolor{linkpink}{RGB}{255,105,180}  
\hypersetup{
    colorlinks=true,
    linkcolor=linkpink,   
    citecolor=linkpink,   
    urlcolor=linkpink,    
    filecolor=linkpink
}

\newcommand{\cmark}{\textcolor{deepgreen}{\ding{51}}}
\newcommand{\xmark}{\textcolor{deepred}{\ding{55}}}

\title{Wh0: Generative World Models as Scalable Sources of Egocentric Human Hand Manipulation Data}

\author{
Yangtao Chen\thanks{These authors contributed equally to this work.}~~$^{1,2}$, Zixuan Chen\footnotemark[1]~~$^{2}$, Peiyang Wang\footnotemark[1]~~$^{2}$, \\ \textbf{Yong-Lu Li\thanks{Corresponding author: Yong-Lu Li (yonglu\_li@sjtu.edu.cn), Jing Huo (huojing@nju.edu.cn).}~~$^{1,3}$, Jing Huo\footnotemark[2]~~$^{2}$, 
 Jieqi Shi$^{2}$, Yang Gao$^{2}$}\\
$^1$Shanghai Innovation Institute, $^2$Nanjing University, $^3$Shanghai Jiaotong University
}

\begin{document}
{\hypersetup{linkcolor=black}
\maketitle}

\begin{abstract}
Scaling dexterous manipulation requires generalization across objects, scenes, and tasks, yet existing data sources face a trade-off between scale and scene/embodiment alignment: teleoperation data is well aligned with robot deployment but expensive to collect; simulation is scalable but limited by the sim-to-real gap; and real egocentric videos scale effectively but remain misaligned with robot deployment. 
We propose \textbf{Wh0}, a framework that uses generative video world models as scalable and controllable sources of egocentric human-hand manipulation data to unlock the manipulation capabilities of pretrained dexterous VLA models. Conditioned on language, objects, and scenes, Wh0 uses a generative world model to produce WM-H, a 50k-episode dataset of egocentric human-object interaction videos. Wh0 then converts the generated videos into robot-trainable supervision through hand motion reconstruction and visual editing. Co-trained with a limited amount of real robot data, WM-H adapts pretrained VLA models to dexterous manipulation deployment. Across 18 real-world dexterous manipulation tasks, compared with a model post-trained only on robot data, Wh0 improves zero-shot success on unseen tasks from 8.3\% to 38.9\%. Ablation studies further show that scalable generation and scene/embodiment alignment are key drivers of performance gains. 
Videos and open-source code can be found on our project website: \href{https://chenyt31.github.io/wh0.github.io/}{https://chenyt31.github.io/wh0.github.io/}.
\end{abstract}

\begin{figure}[ht!]
    \centering
    \includegraphics[width=\linewidth]{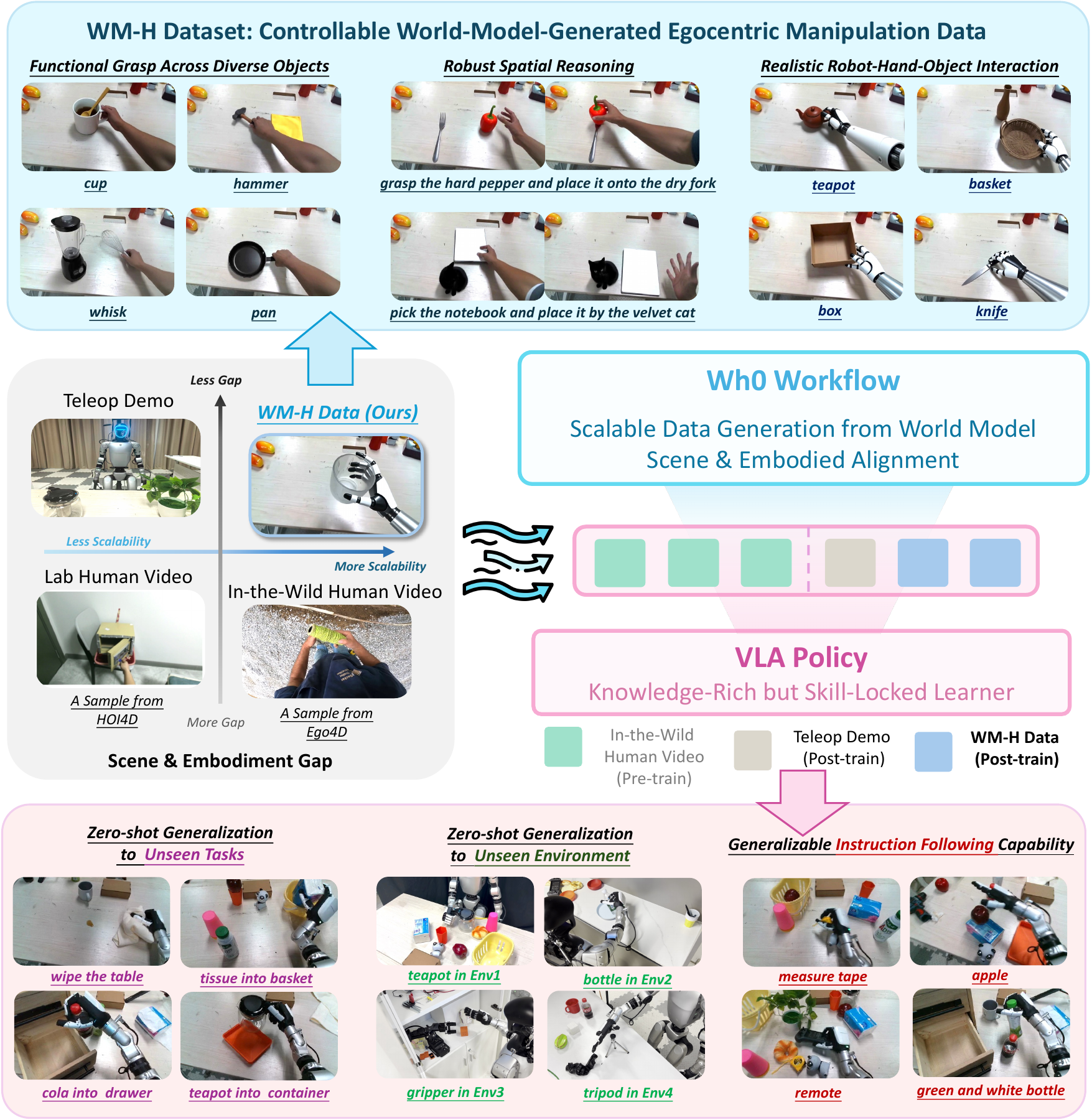}
    \caption{\small \textbf{Overview of Wh0.} \textbf{Top:} WM-H provides world-model-generated egocentric manipulation videos with diverse objects, layouts, and hand-object interactions. \textbf{Middle:} WM-H uniquely combines scale with low scene \& embodiment gap to deployment; Wh0 converts them to robot-trainable supervision and co-trains with limited robot data atop a human-video-pretrained VLA. \textbf{Bottom:} The resulting policy zero-shot generalizes to unseen tasks, environments, and instructions in real-world manipulation.}
    \label{fig:teaser}
\end{figure}

\section{Introduction}

Modern VLA models~\cite{rt-1,rt-2,openvla,pi0} achieve strong generalization by leveraging large-scale data, especially egocentric human videos~\cite{egovla,egodex,li2025VITRA,beingh0,beingh05,beingh07,Maple,lepertphantom}, but still face scene and embodiment misalignment when deployed to dexterous manipulation. Specifically, everyday human environments~\cite{Ego4d,egolive} differ from robot workspaces, leading to a scene gap. Meanwhile, the acting hand in egocentric videos is a human hand rather than a dexterous robot hand~\cite{EgoMimic,zeromimic}, resulting in an embodiment gap.
Existing data sources address only part of this trade-off. Teleoperation aligns with deployment but is expensive and platform-specific~\cite{OXE,droid}. Simulation scales but suffers from a sim-to-real gap~\cite{isaac,sim2real}. Real egocentric video scales but remains misaligned~\cite{Ego4d,epic-kitchens}. An ideal data source should be scalable and deployment-aligned, while minimizing reliance on human data collection.

In this paper, \textit{we formulate generative video world models as scalable, compute-driven sources for synthesizing egocentric human-hand manipulation data.} Their key advantage lies not in precisely simulating robot dynamics, but in generating diverse human-object interaction videos on demand, conditioned on language, objects, and scenes~\cite{wan2025wan,cosmos}. Therefore, unlike prior work that uses world models as environment dynamics simulators~\cite{WM,3d-vla,DexWM,gwm}, robot trajectory video generators~\cite{DreamGen,du2023learning,bharadhwaj2024gen2act}, human-hand generators for downstream retargeting or affordance inference~\cite{liang2024dreamitate, chen2025largevideoplanner,li2025novaflow, kim2025dexterousworldmodels}, or backbones for training future-predictive policies~\cite{dreamzero,motubrain,lingbotva}, we focus on their role as controllable human-hand data generators, where scenes, objects, task types, and embodiment appearance become design variables that scale with compute rather than human labor.

As shown in Fig~\ref{fig:teaser}, we instantiate this idea with \textbf{Wh0}, a framework that uses generative video world models to build egocentric human-hand manipulation data for unlocking dexterous capabilities in pretrained VLA models. Wh0 constructs \textbf{WM-H}, a 50k-sample dataset with egocentric manipulation videos, language instructions, and 3D hand motion annotations, via automated instruction generation, scene- and embodiment-aligned video synthesis, visual editing, and hand motion reconstruction. During post-training, Wh0 co-trains WM-H with limited real teleoperated robot data: robot data supplies deployment-specific constraints, while WM-H bridges human manipulation priors and robot execution through a unified hand action space and embodiment alignment.

On a Unitree G1 humanoid with Inspire dexterous hands, under zero-shot evaluation, the strong VLA baseline VITRA~\cite{li2025VITRA} post-trained only on robot data without test-task demonstrations achieves an 8.3\% success rate, while co-training with WM-H improves it to 38.9\%, a 4.7$\times$ gain. Ablations verify that scene and embodiment alignment, together with data scale, are the primary drivers of Wh0's gains. A separate analysis shows that WM-H exposes pretraining-acquired manipulation capabilities inaccessible with limited robot data alone.


\section{Related Work}

\textbf{Vision-Language-Action Models for Dexterous Manipulation.}
Vision-language-action (VLA) models~\cite{rt-1,rt-2,openvla,pi0,gr00t,pi05,pi07} extend vision-language models to robot control by learning general-purpose policies from large-scale robots and heterogeneous data~\cite{OXE,Bc-z,Bridgedata}. Recent work explores vision-language models for dexterous manipulation, including hierarchical VLA frameworks~\cite{DexGraspVLA} and VLM-based planning systems~\cite{RoboDexVLM}. To reduce reliance on costly dexterous demonstrations of real-robots, prior work uses egocentric human videos~\cite{Ego4d,epic-kitchens}, often with hand reconstruction~\cite{hawer,hawor}, to learn manipulation priors, affordances, transferable hand actions, and policy representations~\cite{affordances,videodex,dexmv,DexCap,EgoMimic,egodex,egovla,li2025VITRA,egoscale,beingh0,pihuman}. However, passive human videos are difficult to control and remain misaligned with robot deployment. We instead construct egocentric human-hand data through controllable generation for deployment-aligned VLA post-training.

\textbf{World Models for Robotic Manipulation.}
World models have been widely used in robotics as dynamics models, visual predictors, and policy-learning substrates~\cite{WM,3d-vla,gwm}. Recent video-generation approaches use world models to synthesize future interaction videos and convert them into robot supervision, such as trajectories, point tracks, object poses, or 3D flows~\cite{liang2024dreamitate,patel2025robotic,li2025novaflow,dharmarajan2025dream2flow,bharadhwaj2024gen2act,DexWM,dreamzero}. 
As general-purpose data engines, Cosmos~\cite{cosmos} and GigaWorld~\cite{GigaWorld-0} treat world models as platforms for robotics.
More related to dexterous manipulation, Large Video Planner generates human-hand or robot-gripper video plans for action extraction and retargeting~\cite{chen2025largevideoplanner}, while Dexterous World Models~\cite{kim2025dexterousworldmodels} and DexWM~\cite{DexWM} model hand-conditioned interaction dynamics from egocentric hand motions or human videos. Wh0 instead treats generated human-hand videos as post-training data, co-trained with limited real robot demonstrations to improve the transfer of manipulation capabilities from human-video pretraining.


\section{WM-H Dataset Construction via Controllable Video Synthesis}
\label{sec:wmh}

\begin{figure}[ht]
    \centering
    \includegraphics[width=\linewidth]{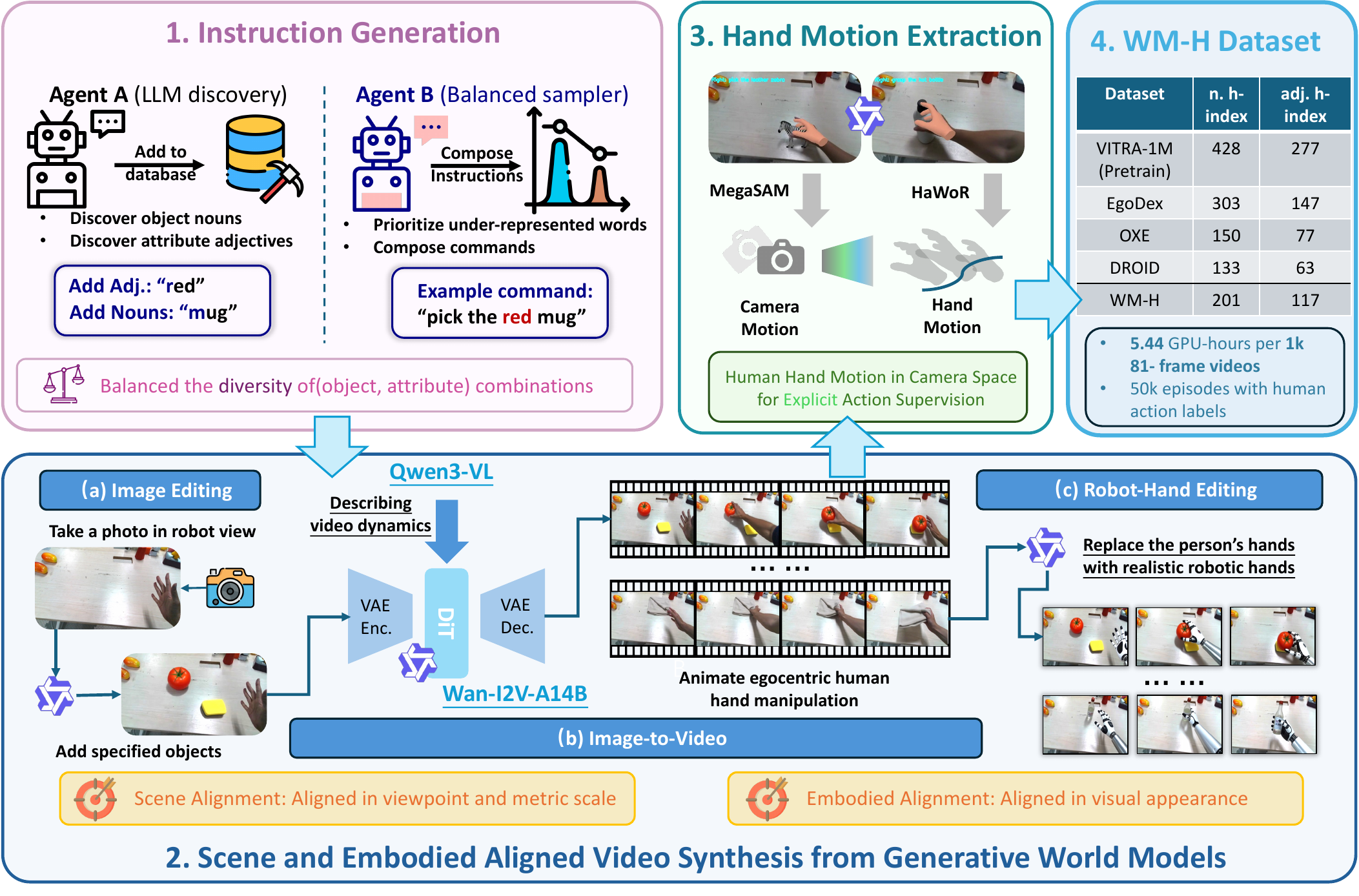}
    \caption{\textbf{\small The overview of WM-H data synthesis pipeline.}}
    \label{fig:pipeline}
\end{figure}

We use video world models as controllable generators to produce egocentric human-hand manipulation data. Conditioned on language instructions, object specifications, and scene layouts, the model synthesizes diverse human-object interaction videos. The scalability of the pipeline mainly depends on GPU compute rather than human labor. As shown in Fig.~\ref{fig:pipeline}, the pipeline consists of instruction generation, scene and embodiment aligned video synthesis, and hand motion extraction, ultimately producing \textbf{WM-H}: a 50k-sample egocentric video dataset with 3D hand pose annotations and language task descriptions. The generation cost is approximately 5.44 GPU-hours per 1k videos. We measure instruction diversity using noun and adjective h-indexes~\cite{li2025VITRA}, where an h-index of $h$ means that at least $h$ words each appear in at least $h$ samples. Despite being much smaller than large-scale human video datasets~\cite{li2025VITRA, egodex, OXE, droid}, WM-H achieves broad manipulation-relevant coverage, with a noun h-index of 201 and an adjective h-index of 117 across pick, place, and grasp instructions.

\textbf{Instruction Generation for Balanced Diversity.}
A key insight behind our instruction generation pipeline is that a diverse manipulation dataset should contain not only many distinct words, but also sufficient coverage of each word~\cite{li2025VITRA}. Simply increasing vocabulary size is insufficient if most words appear only rarely. To balance breadth and frequency, we employ a dual-agent system as shown in Fig.~\ref{fig:pipeline}(1). The first agent uses a large language model to continually discover object nouns and attribute adjectives related to graspable items, expanding the candidate vocabulary pool. The second agent preferentially samples under-represented words and assembles natural language commands using structured templates such as \textit{pick the \{adj\} \{noun\}}. A database tracks word usage frequencies and filters duplicate instructions, enabling systematic and balanced coverage of the space of (object, attribute) combinations.

\textbf{Video Synthesis with Scene and Embodiment Alignment.}
Given a generated instruction, we synthesize an egocentric manipulation video through the three-stage pipeline shown in Fig.~\ref{fig:pipeline}. The pipeline aims to preserve the natural hand-object interaction patterns of video world models while reducing the visual gap to robot deployment:
\textbf{(a) Scene image capture and editing.}
We first capture background images in the target robot workspace for scene alignment. The images are collected by the deployment camera, with the same viewpoint and resolution as the policy input. During capture, we place a human hand in the target interaction region as a scale anchor, providing references for hand size, object scale, and camera-to-workspace distance. Based on the aligned background, we use Qwen-Image-Edit~\cite{qwen-image} to insert the specified objects into the scene, producing the initial frame for video synthesis.
\textbf{(b) Image-to-video generation.}
We use Wan-I2V-A14B~\cite{wan2025wan} to animate the edited image and generate an egocentric human-hand manipulation video conditioned on the language instruction. To improve the correctness of the generated motion, we use Qwen3-VL~\cite{qwen3vl} to generate a description of the expected hand-object state changes and append it to the video prompt. To enable efficient large-scale synthesis, we adopt LightX2V LoRA adapters, reducing video generation to four inference steps.
\textbf{(c) Robot-hand editing.}
For embodiment alignment, we treat a robotic dexterous hand as a hand entity with a different visual appearance. We use Qwen-Image-Edit~\cite{qwen-image} to replace the human hand with a realistic robot hand on selected frames, while preserving the original pose, position, object motion, and scene composition. This renders the same manipulation trajectory with both human and robot hand appearances, encouraging the policy to focus on action semantics rather than executor identity. We ablate this design in Section~\ref{sec:experiments:why}.
For the evaluation of video quality and the prompts involved in the pipeline, please refer to Appendix~\ref{sec:APP-WM-H}.

\textbf{Motion Extraction for Explicit Action Supervision.}
Follow~\cite{li2025VITRA}, we extract 3D hand poses from synthesized human-hand videos as action labels. Generating human rather than robot manipulation videos is important here: human hand reconstruction is comparatively mature, making it feasible to recover reliable motion supervision from generated videos at scale.
For each video, we use HaWoR~\cite{hawor} to reconstruct the hand motion. The method first detects hands in each frame and then regresses MANO parameters~\cite{MANO} together with the wrist pose. We keep the wrist pose in camera space and the articulated hand pose in the MANO parameter space. When needed, camera tracking from MegaSAM is used to associate these per-frame predictions with the video camera trajectory.

\section{Wh0: Policy Learning with Human-Robot Alignment}
\label{sec:policy}

\begin{figure}[ht]
    \centering
    \includegraphics[width=\linewidth]{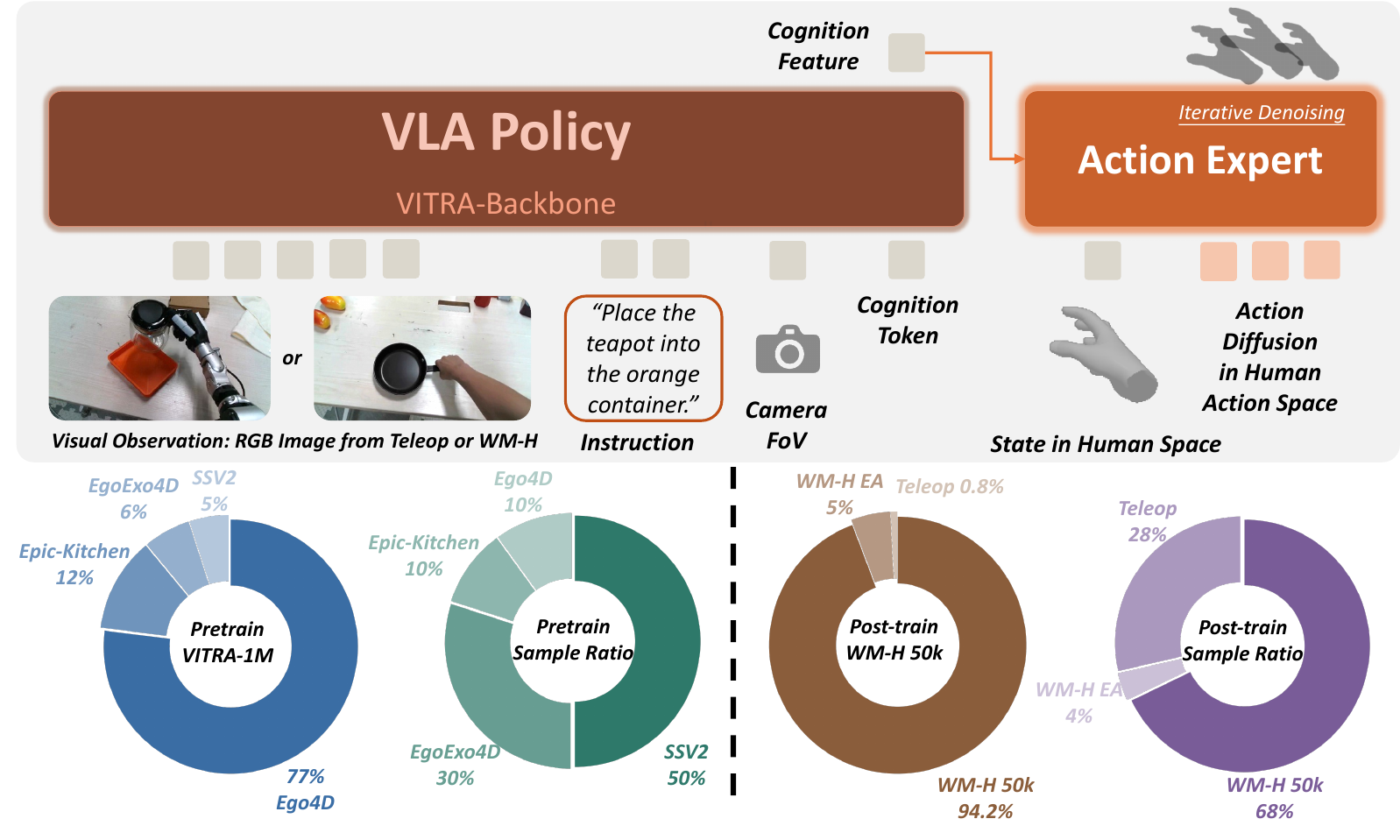}
    \caption{\small \textbf{ Policy architecture and data composition.}
    \textbf{Top:} A VITRA-style policy denoises actions in the unified MANO space, conditioned on PaliGemma cognition features, FoV, and current hand state.
    \textbf{Bottom:} Pretraining mixture (VITRA-1M, Ego4D-dominant) and post-training mixtures: Wh0 uses 28\% teleop and 68\% WM-H, heavily oversampling robot data per-sample given 400 teleop vs.\ 50k WM-H samples.}
    \label{fig:recipe}
\end{figure}

\paragraph{Policy Architecture}
We adopt a VITRA-style vision-language-action policy~\cite{li2025VITRA}: a PaliGemma~\cite{beyer2024paligemma} vision-language backbone encodes the current observation and language instruction (with FoV as an auxiliary token for viewpoint cues), and the resulting feature conditions a diffusion-based action decoder that predicts future hand motions from the current hand state.

The action space is defined in the camera coordinate frame of the current observation $o_t$:
\begin{equation}
a_t = [\Delta t^l, \Delta r^l, \theta_h^l, \Delta t^r, \Delta r^r, \theta_h^r] \in \mathbb{R}^{102},
\end{equation}
where $\Delta t, \Delta r \in \mathbb{R}^3$ denote the relative wrist translation and rotation (Euler angles) between consecutive frames, and $\theta_h \in \mathbb{R}^{15 \times 3}$ encodes the joint rotations of a 15-DoF MANO~\cite{MANO} hand model in its local frames. Superscripts $l, r$ denote left and right hands; we focus on right-hand manipulation in this work.
We retarget robot joints to MANO and reuse the per-joint normalization parameters precomputed by VITRA~\cite{li2025VITRA} from large-scale human videos. This avoids robot-specific normalization from limited data and keeps robot actions aligned with the human action space.

The diffusion action decoder is trained with a noise-prediction MSE loss:
\begin{equation}
\mathcal{L}_{\mathrm{MSE}}
=
\mathbb{E}_{\epsilon \sim \mathcal{N}(0, 1),\, i}
\left[
\left\|
\hat{\epsilon}_i - \epsilon
\right\|_2^2
\right],
\end{equation}
where $\hat{\epsilon}_i$ is the predicted noise at diffusion step $i$.

\paragraph{Training Strategy}
The policy is initialized from VITRA~\cite{li2025VITRA}, pretrained on Ego4D~\cite{Ego4d}, Epic-Kitchens~\cite{epic-kitchens}, EgoExo4D~\cite{grauman2024egoexo4d}, and Something-Something-V2~\cite{goyal2017something} (per-dataset ratios in Fig.~\ref{fig:recipe}), giving the model strong priors on human hand-object interaction.
We then co-finetune on a 125:1 mixture of 50k WM-H samples and 400 real teleoperated robot demonstrations. Each batch draws 28\% from teleop, 68\% from WM-H, and 4\% from WM-H EA (WM-H frames after robot-hand editing for embodiment alignment). This oversamples the scarce robot data, providing stable embodiment-specific signals, while WM-H supplies the visual and semantic diversity needed for generalization. We freeze the vision encoder and update the remaining backbone together with the diffusion action decoder, using learning rate $1\times10^{-5}$ and VLM weight decay $0.1$. For additional details on training, please refer to Appendix~\ref{sec:APP-Policy}.

\section{Experiments}

\begin{figure}
    \centering
\includegraphics[width=\linewidth]{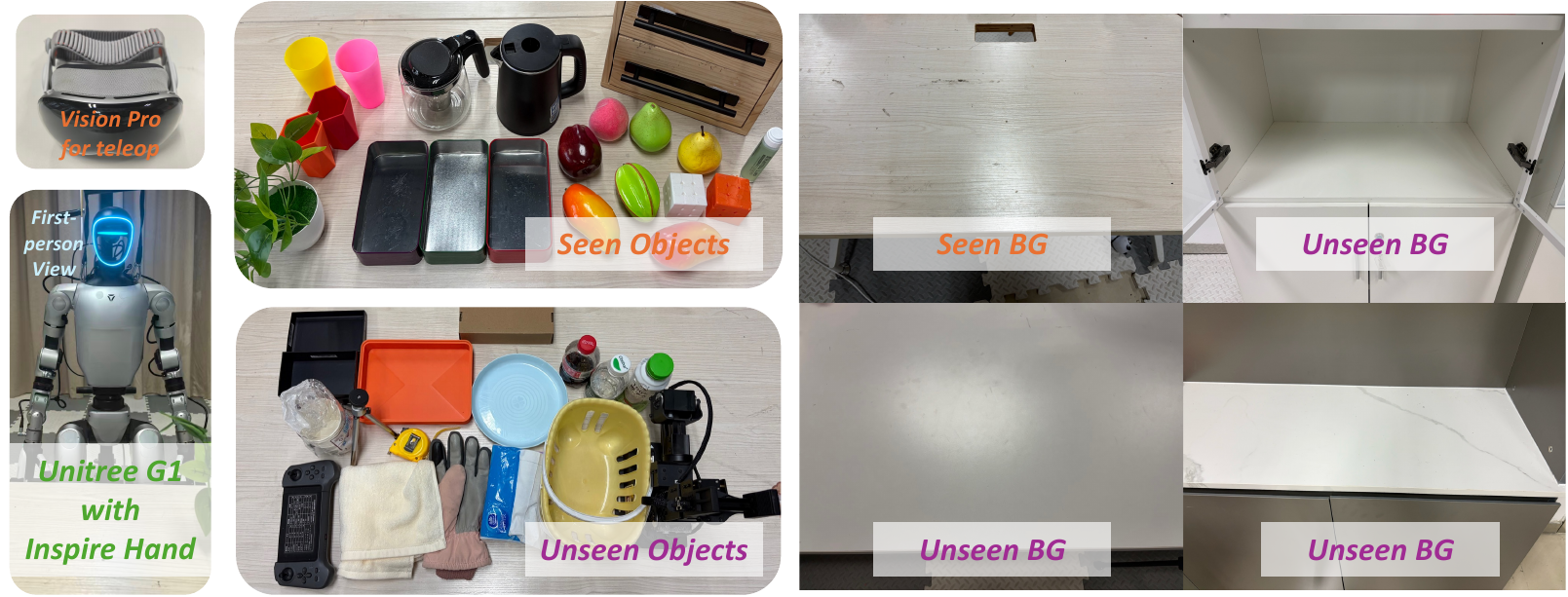}
    \caption{\small \textbf{Real-world evaluation setup.} Unitree G1 with Inspire hands and a head-mounted egocentric camera (teleop via Vision Pro); evaluation spans seen/unseen objects and one seen plus three unseen backgrounds.}
    \label{fig:exp-setup}
\end{figure}

\begin{figure}
    \centering
    \includegraphics[width=\linewidth]{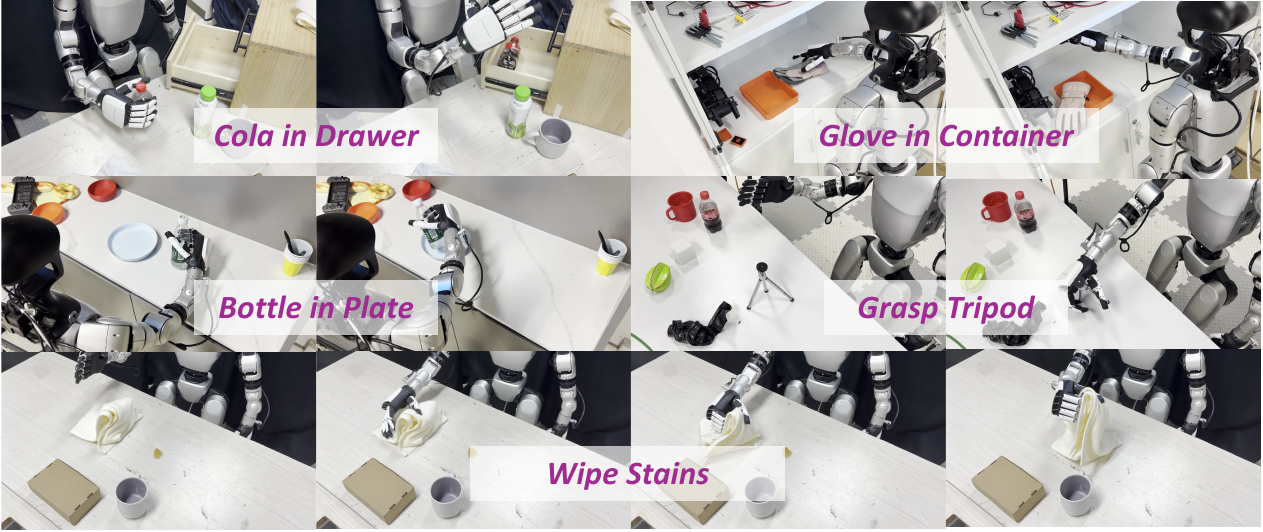}
    \caption{\small \textbf{Zero-shot rollouts on real-world dexterous tasks}, including container-aware placement, small-object grasping, and tool use. None of these object, container, or task combinations appear in the training set.}
    \label{fig:Q1}
\end{figure}

We evaluate Wh0, a post-training framework that uses world-model-generated human-hand data to improve VLA policies for dexterous manipulation, through three questions:
\textbf{(Q1) Real-World Performance}: Does Wh0 improve zero-shot real-world dexterous manipulation over strong VLA baselines?
\textbf{(Q2) Ablations and Analysis}: Which properties of WM-H drive dexterous generalization?
\textbf{(Q3) Pretraining Priors}: Does WM-H unlock pretrained human manipulation priors?

\subsection{Experimental Setup}

\textbf{(1) Robot platform.} We evaluate Wh0 on a Unitree G1 humanoid with Inspire dexterous hands; all methods share the same hardware, camera, and control interface. \textbf{(2) Training data.} We collect 400 expert trajectories via Apple Vision Pro on the seen tasks and backgrounds in Fig.~\ref{fig:exp-setup}, mainly pick-and-place over the shown objects. Unless otherwise specified, all policies train on these trajectories; mixed-training setups are explicitly noted. 
\textbf{(3) Evaluation protocol.} The benchmark contains 18 tasks across four scenes (grasping, placing, object-specific interactions), each evaluated over 20 trials with randomized object poses and scenes; success rate is the primary metric. All policies are evaluated \textbf{zero-shot} without task-specific demonstrations.

\subsection{Q1: Does Wh0 Improve Real-World Zero-Shot Dexterous Manipulation?}

\begin{table}[t]
\centering
\scriptsize
\setlength{\tabcolsep}{3.8pt}
\renewcommand{\arraystretch}{1.04}
\resizebox{\linewidth}{!}{%
\begin{tabular}{@{}lcccc@{}}
\toprule
\multirow{2}{*}{\textbf{Method}} &
\multicolumn{3}{c}{\textbf{Training Setup}} &
\multirow{2}{*}{\textbf{Success Rate (\%) $\uparrow$}} \\
\cmidrule(lr){2-4}
&
\textbf{Pretraining} &
\textbf{Adaptation Data} &
\textbf{Strategy} &
\\
\midrule
$\pi_{0.5}$              & Robot & Teleop              & FT    & $7.78_{\scriptscriptstyle \pm 15.6}$ \\
VITRA                    & Human & Teleop              & FT    & $8.3_{\scriptscriptstyle \pm 8.6}$ \\
VITRA Real Version       & Human & Teleop + Real Ego   & Co-FT & $21.4_{\scriptscriptstyle \pm 23.4}$ \\
\midrule
\textbf{Wh0}             & Human & Teleop + WM-H       & Co-FT & $\mathbf{38.9}_{\scriptscriptstyle \pm 19.8}$ \\
\bottomrule
\end{tabular}}
\caption{\small \textbf{Real-world dexterous manipulation performance.}
We compare different pretraining sources and adaptation data under the same real-robot evaluation protocol.
FT denotes fine-tuning on a single adaptation source, while Co-FT denotes joint fine-tuning on multiple data sources.}
\label{tab:main_results}
\end{table}

We compare \textbf{Wh0} against representative VLA adaptation baselines that differ in pretraining source and adaptation data:
$\boldsymbol{\pi}_{0.5}$~\cite{black2025pi}, a robot-data-pretrained policy adapted with teleoperated demonstrations;
\textbf{VITRA}~\cite{li2025VITRA}, a human-video-pretrained VLA model adapted with the same teleoperated demonstrations;
and \textbf{VITRA Real Version}, which replaces WM-H with real egocentric human-hand manipulation videos from the HOI4D dataset~\cite{HOI4D} during co-training.
The conventional paradigm adapts pretrained policies via robot post-training alone, but $\pi_{0.5}$ and VITRA show limited instruction-following in our experiments, suggesting that post-training on limited robot data overfits to seen tasks and weakens pretraining generalization.
By contrast, co-training with additional human data, either lab-collected videos or WM-H, substantially improves zero-shot success. Moreover, Fig.~\ref{fig:Q1} qualitatively shows more deployment-aligned rollouts from Wh0 than from real egocentric videos, suggesting that world-model-generated data provides more suitable supervision for robot execution.
For further experimental results, please refer to Appendix~\ref{sec:APP-EVAL}.

\subsection{Q2: Which Properties of WM-H Matter?}
\label{sec:experiments:why}

\begin{wraptable}{r}{0.55\textwidth}
\centering
\small
\setlength{\tabcolsep}{4pt}
\renewcommand{\arraystretch}{0.9}
\begin{tabular}{lccc}
\toprule
\textbf{Variant} & \makecell{\textbf{HO} \\ \textbf{(human)} $\uparrow$} & \makecell{\textbf{HO} \\ \textbf{(robot)} $\uparrow$} & \makecell{\textbf{Task} \\ \textbf{Succ.} $\uparrow$} \\
\midrule
\textit{No model} & $18.9_{\scriptscriptstyle \pm 2.8}$ & $18.9_{\scriptscriptstyle \pm 2.8}$ & -- \\
Teleop only & $16.2_{\scriptscriptstyle \pm 3.3}$ & $16.2_{\scriptscriptstyle \pm 3.3}$ & $8.3_{\scriptscriptstyle \pm 8.6}$ \\
\midrule
\multicolumn{4}{l}{\textit{Alignment ablations}} \\
w/o scene align. & $14.9_{\scriptscriptstyle \pm 2.7}$ & $14.3_{\scriptscriptstyle \pm 2.5}$ & $20.0_{\scriptscriptstyle \pm 24.7}$ \\
w/o emb. align. & $\mathbf{10.2}_{\scriptscriptstyle \pm 2.5}$ & $13.8_{\scriptscriptstyle \pm 3.6}$ & $34.7_{\scriptscriptstyle \pm 18.0}$ \\
\midrule
\multicolumn{4}{l}{\textit{Scale ablations}} \\
WM-H 5k & $11.9_{\scriptscriptstyle \pm 2.8}$ & $10.5_{\scriptscriptstyle \pm 3.2}$ & $27.8_{\scriptscriptstyle \pm 21.8}$ \\
WM-H 25k & $11.4_{\scriptscriptstyle \pm 2.5}$ & $9.9_{\scriptscriptstyle \pm 2.6}$ & $32.5_{\scriptscriptstyle \pm 23.5}$ \\
\textbf{Wh0 (50k)} & $10.6_{\scriptscriptstyle \pm 2.0}$ & $\mathbf{9.6}_{\scriptscriptstyle \pm 1.8}$ & $\mathbf{38.9}_{\scriptscriptstyle \pm 19.8}$ \\
\bottomrule
\end{tabular}
\caption{\small \textbf{Ablation Study:} effects of deployment alignment and WM-H scale on robot-object grounding and task success. HO = Hand-Object Distance (cm).}
\label{tab:ablation}
\end{wraptable}

\begin{figure}[t!]
    \centering
    \includegraphics[width=\linewidth]{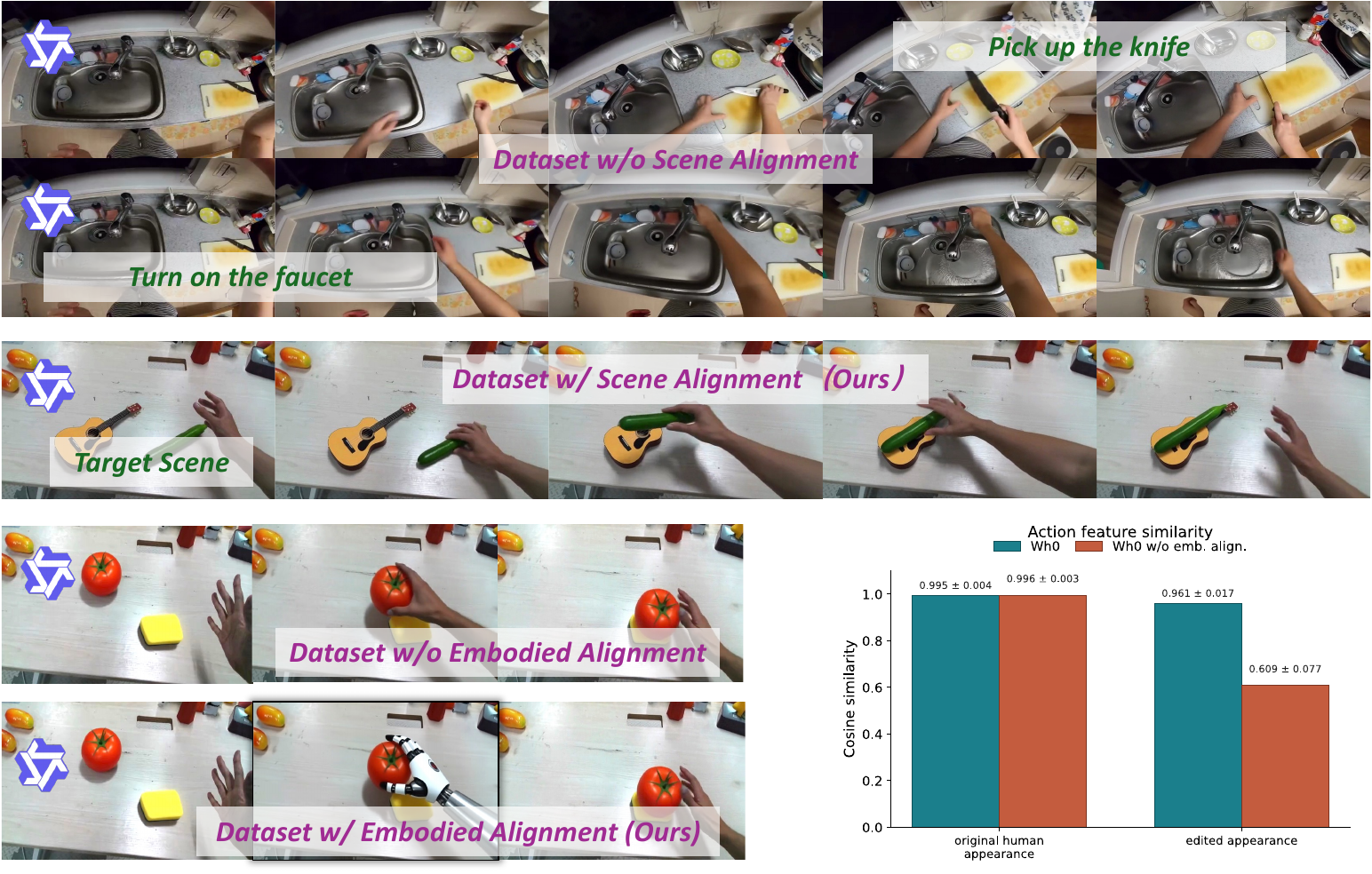}
  \caption{\small \textbf{Effect of scene and embodiment alignment.} \textbf{Top:} Without scene alignment, generated videos drift from the target workspace (left); with it (\textit{ours}), they stay anchored. \textbf{Middle:} Embodiment alignment edits selected frames to a robot hand while preserving pose and motion. \textbf{Right:} Action-feature cosine similarity under original vs.\ edited appearance. 
  }
    \label{fig:ablation}
\end{figure}

We analyze the factors behind the gains in Table~\ref{tab:main_results} through two ablations: removing scene and embodiment alignment from WM-H, and varying the number of WM-H samples used for co-training.
As shown in Table~\ref{tab:ablation} and Fig.~\ref{fig:ablation}, we evaluate these factors with three metrics: Hand-Object Distance human, which measures grounding on original human-hand videos; Hand-Object Distance robot, which measures grounding stability after robot-hand appearance editing; and real-world task success. \textbf{Without scene alignment}, WM-H still provides hand-object interaction patterns, but the scene distribution and viewpoint no longer match deployment, limiting both grounding and real-world success. \textbf{Without embodiment alignment}, the model performs well under the human-hand appearance, but degrades under the robot-hand appearance and achieves lower task success than full Wh0, indicating that its grounding does not reliably transfer to robot embodiment. Fig.~\ref{fig:ablation} further shows that embodiment alignment keeps action features more stable under appearance changes. Finally, increasing WM-H scale consistently improves both grounding metrics and real-world success, showing that aligned data becomes more effective as it scales.

\subsection{Q3: Does WM-H Unlock Pretrained Human Manipulation Priors?}
\label{sec:experiments:unlocking}

\begin{table}[ht]
\centering
\setlength{\tabcolsep}{3.8pt}
\renewcommand{\arraystretch}{1.03}
\resizebox{\linewidth}{!}{%
\begin{tabular}{@{}lccc|cc@{}}
\toprule
\multirow{2}{*}{\textbf{Variant}} &
\multicolumn{3}{c|}{\textbf{Training Sources}} &
\multicolumn{2}{c}{\textbf{Evaluation Metrics}} \\
\cmidrule(lr){2-4} \cmidrule(l){5-6}
&
\textbf{Human Pretrain} &
\textbf{Teleop} &
\textbf{WM-H} &
\textbf{Hand-Object Dist. (cm) $\downarrow$} &
\textbf{Task Success $\uparrow$} \\
\midrule
\textit{No model (initial pose)} & -- & -- & -- & $18.9_{\scriptscriptstyle \pm 2.8}$ & -- \\
\textit{PaliGemma pretrain}, Teleop & \xmark & \cmark & \xmark & $14.3_{\scriptscriptstyle \pm 2.0}$ & $0.8_{\scriptscriptstyle \pm 2.6}$ \\
\textit{PaliGemma pretrain}, Teleop + WM-H & \xmark & \cmark & \cmark & $12.7_{\scriptscriptstyle \pm 1.4}$ & $0.6_{\scriptscriptstyle \pm 1.6}$ \\
\textit{Human pretrain} & \cmark & \xmark & \xmark & $13.1_{\scriptscriptstyle \pm 1.8}$ & $0.0$ \\
\textit{Human pretrain}, Teleop & \cmark & \cmark & \xmark & $16.2_{\scriptscriptstyle \pm 3.3}$ & $8.3_{\scriptscriptstyle \pm 8.6}$ \\
\midrule
\textbf{Wh0} & \cmark & \cmark & \cmark & $\mathbf{10.6}_{\scriptscriptstyle \pm 2.0}$ & $\mathbf{38.9}_{\scriptscriptstyle \pm 19.8}$ \\
\bottomrule
\end{tabular}}
\caption{\small \textbf{Unlocking pretraining priors:} effects of human-video pretraining, teleop demonstrations, and WM-H co-training. The strongest result combines all three.}
\label{tab:unlocking}
\end{table}

Table~\ref{tab:unlocking} studies whether WM-H helps leverage human-video pretraining. Human pretraining alone achieves low Hand-Object Distance but zero task success, suggesting general human hand-motion priors, consistent with the evaluation in~\cite{li2025VITRA}, but these priors are not directly deployable on the robot. Without human pretraining, teleoperation data and WM-H still yield near-zero success. In contrast, combining human pretraining, teleoperation, and WM-H achieves the best grounding and success. These results are consistent with WM-H helping activate and align pretrained human manipulation priors, rather than learning dexterous skills from scratch.

\section{Conclusion and Limitations}
We presented \textbf{Wh0}, a framework that uses generative video world models to build egocentric human-hand manipulation data for VLA post-training. By constructing the 50k-sample \textbf{WM-H} dataset and co-training it with limited real robot demonstrations, Wh0 improves zero-shot dexterous manipulation on a Unitree G1 humanoid with Inspire hands, increasing a strong VITRA baseline from 8.3\% to 38.9\% success, a 4.7$\times$ improvement. Ablations show that scene alignment, embodiment alignment, and data scale all contribute to the gains, while further analysis indicates that WM-H mainly unlocks human-video-pretrained manipulation priors rather than learning dexterous skills from scratch. These results support world-model-generated human-hand videos as scalable, deployment-aligned post-training data for dexterous VLA policies.

Wh0 remains limited by video generation quality, hand reconstruction accuracy, human-robot morphology mismatch, dependence on strong pretraining, and task scope. The generator may produce physically implausible interactions, unexpected objects, or inconsistent long-horizon videos. Hand-object occlusions can degrade reconstructed finger poses and introduce noisy supervision, while the robot hand's larger size can unintentionally disturb objects during execution. WM-H also provides little benefit without a human-video-pretrained backbone, indicating that it complements rather than replaces large-scale pretraining. Finally, our experiments focus on single-arm pick-and-place manipulation; extending Wh0 to bimanual, tool-use, and longer-horizon tasks remains future work.


\bibliography{main-ref}  

\clearpage
\appendix
\section*{Appendix}
\section{WM-H Dataset Construction}
\label{sec:APP-WM-H}

\subsection{Instruction Generation and Prompt Design}

The pipeline is implemented as a database-driven iterative process. The database tracks nouns, adjectives, instructions, and word frequencies. At each iteration, low-frequency words are prioritized for instruction assembly, and vocabulary expansion is triggered only when all existing words have reached minimum usage thresholds. Newly generated words are parsed from a JSON output returned by the LLM, filtered, and inserted into the database, while duplicate instructions are rejected. As shown in Tab.~\ref{box:vocab_prompt}, the vocabulary-expansion prompt includes: \texttt{\{num\_nouns\}} and \texttt{\{num\_adjectives\}} to specify the number of new words; \texttt{\{existing\_nouns\}} and \texttt{\{existing\_adjectives\}} as a compact subset to avoid duplicates; and \texttt{\{diversity\_hint\}} suggesting a scene category (e.g., kitchen, office) to encourage diverse word types.

\begin{tcolorbox}[
    colback=gray!5,
    colframe=gray!60,
    title=Vocabulary Expansion Prompt,
    label={box:vocab_prompt}
]
\small
\texttt{Generate \{num\_nouns\} new nouns and \{num\_adjectives\} new adjectives for graspable manipulation objects.}\\
\texttt{Existing nouns: \{existing\_nouns\}}\\
\texttt{Existing adjectives: \{existing\_adjectives\}}\\
\texttt{Diversity hint: \{diversity\_hint\}}\\
\texttt{Return only a JSON object:}\\
\texttt{\{"nouns": [...], "adjectives": [...]\}}
\end{tcolorbox}

\subsection{Scene-Aligned Image Editing}
\label{app:scene_aligned_editing}

We collect desktop/workspace images using the same deployment camera, viewpoint, and input resolution as the robot policy. A human hand is placed in the target interaction area during capture to provide relative scale cues for hand size, object size, and camera distance. For each instruction, we randomly select a workspace image and insert the required objects with Qwen-Image-Edit.

To localize editing, we draw a rectangular guide region on the image. The region covers a fixed proportion of the image and its center is randomly sampled within the reachable workspace area. For multiple objects, guide regions are sampled sequentially while avoiding overlap. The rectangle is used only to indicate the desired insertion area and is removed in the final edited image. We use a small number of editing steps and a low CFG scale with Lightning LoRA acceleration. The edited image is then used as the initial frame for video synthesis. The prompt template is shown in Tab.~\ref{tab:scene_edit_prompt}.

\begin{tcolorbox}[colback=gray!5,colframe=gray!60,title=Scene-Aligned Image Editing Prompt,label={tab:scene_edit_prompt}]
\small
Add \{object\} inside the rectangle drawn on the image. The rectangle is only a guide. The final image must not show the rectangle. Keep the original desktop layout, lighting, camera perspective, background, and existing objects unchanged.
\end{tcolorbox}

\subsection{Image-to-Video Generation}
\label{app:i2v_generation}

Given the edited scene image, we use it as the initial frame for image-to-video generation. The video prompt is constructed from the task instruction, the egocentric camera setting, the acting hand, and an optional dynamics description generated by Qwen3-VL. The Qwen3-VL prompt asks the model to infer the expected hand-object motion and temporal evolution from the edited image and language instruction, as shown in Tab.~\ref{tab:qwen3vl_dynamics_prompt}. The resulting description is appended to the final video prompt in Tab.~\ref{tab:i2v_prompt}.

We keep the camera fixed and specify a first-person top-down view with a single acting hand entering from the lower part of the frame. The generated video focuses on completing the instructed manipulation, and the hand is encouraged to stop after the action is completed. We use LightX2V LoRA acceleration with a small number of inference steps and a low CFG scale for efficient large-scale synthesis.

\begin{tcolorbox}[colback=gray!5,colframe=gray!60,title=Qwen3-VL Video Dynamics Prompt,label={tab:qwen3vl_dynamics_prompt}]
\small
The image shows a scene, and a person's \{LEFT/RIGHT\} hand will perform the following action in the frame: \{instruction\}. 
The task is completed using only the \{LEFT/RIGHT\} hand. 
Please carefully observe the objects and layout in the image, and based on this action, describe in detail the video dynamics that should occur during the action. 
Focus on the hand motion, object motion, hand-object interaction, and the temporal order of the manipulation. 
Make sure to mention that the \{LEFT/RIGHT\} hand is performing the action. 
Please describe the video dynamics in English, output the description text directly, do not add any other content. Be specific.
\end{tcolorbox}

\begin{tcolorbox}[colback=gray!5,colframe=gray!60,title=Image-to-Video Generation Prompt,label={tab:i2v_prompt}]
\small
4k high definition, fixed camera, first-person top-down view, showing a human \{left/right\} hand and objects on the desktop. 
The \{left/right\} hand extends from the lower part of the frame. 
A person's \{LEFT/RIGHT\} hand performs the following action in the frame: \{instruction\}. 
Only the \{left/right\} hand is used to complete this task. 
\{video dynamics description from Qwen3-VL\} 
After the action is completed, the hand stops, with no extra movements.
\end{tcolorbox}

\subsection{Embodiment-Aligned Image Editing}
\label{app:embodiment_aligned_editing}

To improve embodiment alignment, we edit selected frames from the generated human-hand videos and replace the human hand with a realistic robot hand while preserving the original manipulation content. In practice, we sparsely sample frames at a fixed interval with a fixed offset, and apply Qwen-Image-Edit to each selected frame independently. The editing is restricted to the hand appearance: we keep the original hand pose, position, scale, scene background, and overall composition unchanged, so that the edited frames remain consistent with the original manipulation trajectory.

We use Qwen-Image-Edit with Lightning LoRA acceleration, using a small number of editing steps and a low CFG scale for efficient processing. The target robot hand appearance is specified as a realistic dexterous hand with a white back shell, black palm, black fingertips, and a silver robotic forearm. To suppress cartoon-like or synthetic rendering artifacts, we additionally use a negative prompt emphasizing realism. The prompt templates are shown in Tab.~\ref{tab:robot_hand_edit_prompt}.

\begin{tcolorbox}[colback=gray!5,colframe=gray!60,title=Embodiment-Aligned Image Editing Prompt,label={tab:robot_hand_edit_prompt}]
\small
\textbf{Positive prompt:} Replace the human hand in the image with a realistic robot hand, with a white back shell, black palm, black fingertips, and a silver robotic forearm. Only change the hand appearance and material. Keep the original hand pose, position, proportion, background, and composition unchanged.

\vspace{0.5em}
\textbf{Negative prompt:} cartoon, anime, comic, toy hand, plastic hand, CGI look, stylized, 3d render, illustration, low poly, fake robot hand, deformed fingers, extra fingers, blurry, low quality
\end{tcolorbox}

\subsection{Quality Evaluation}

\begin{figure}[t!]
    \centering
    \includegraphics[width=1.0\linewidth]{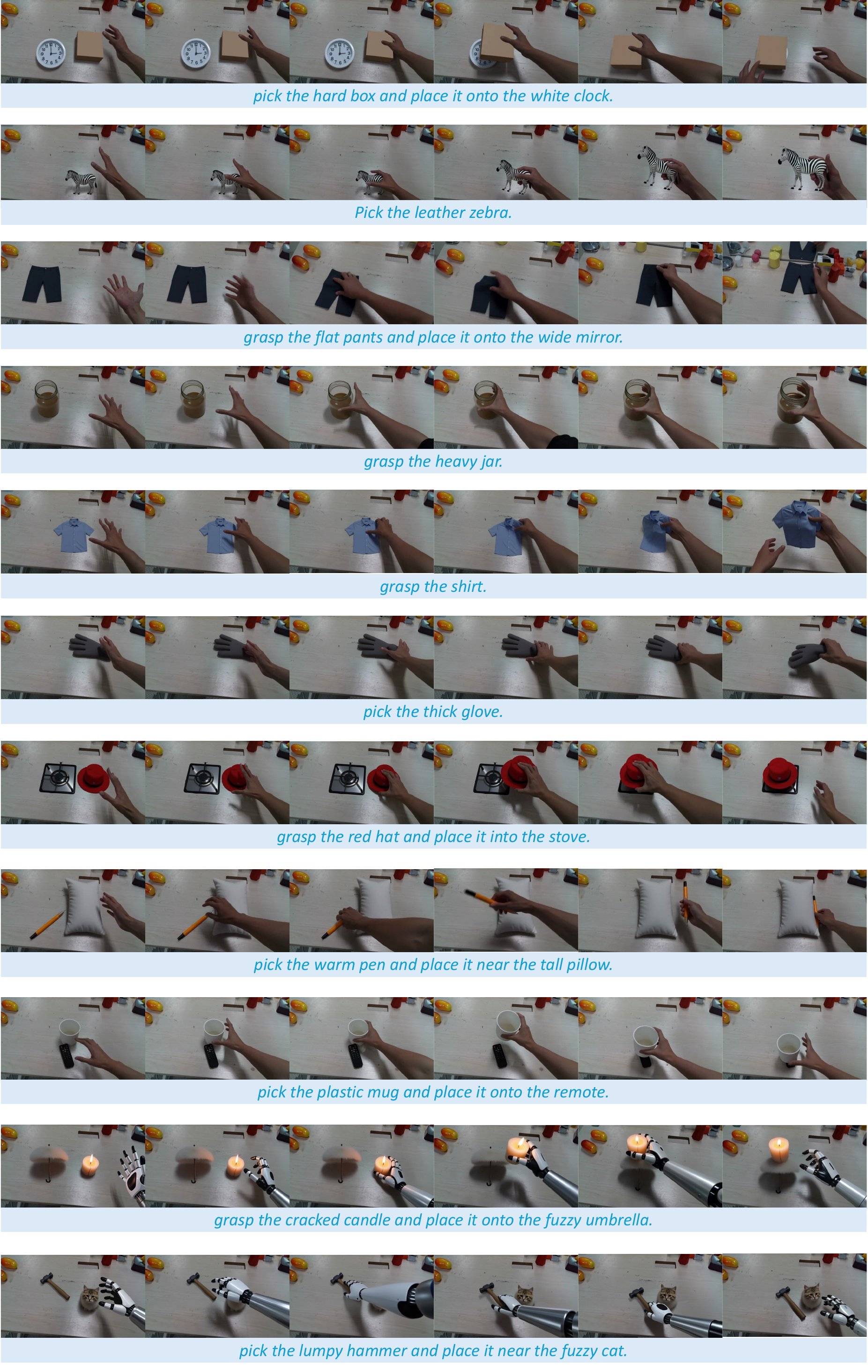}
  \caption{\small Qualitative Visualization of WM-H
  }
    \label{fig:wmh-tasks-1}
\end{figure}

Figure~\ref{fig:wmh-tasks-1} presents a qualitative visualization of egocentric hand-manipulation videos in the WM-H dataset across a variety of objects and tasks. Each row demonstrates a different manipulation, showing how human or robot hands interact with objects in realistic scenes. The figure highlights the diversity of grasp types, object shapes, and placements captured in WM-H, illustrating its potential to provide rich training supervision for vision-language-action models.

We conducted a user study with 72 AI practitioners. As summarized in Table~\ref{tab:user-study}, analysis of judgment cues in Part A shows that participants primarily relied on physics- and layout-related cues (45.6\%) to assess video realism, followed by hand appearance and motion (34.9\%), while low-level visual artifacts such as flickering or distortion accounted for only 19.4\%. This indicates that even evaluators with AI expertise focus more on semantic consistency, physical plausibility, and hand-object interaction rather than solely on temporal smoothness or local visual artifacts. Under these higher-level evaluation criteria, 37.7\% of AI-generated videos (134/355 trials) were perceived as real recordings, demonstrating that a substantial portion of synthetic samples already exhibits strong realism in scene layout, physical plausibility, and manipulation performance.

In Part B, each participant rated 20 synthetic hand-manipulation videos randomly sampled from the WM-H dataset (with randomized sampling and presentation order per session) on five quality dimensions using a 5-point Likert scale. Assuming a real-video ceiling of 5.0 (not directly measured in this study), the average scores for the synthetic videos were $3.97 \pm 1.22$ (object correctness), $4.18 \pm 1.09$ (instruction alignment), $3.95 \pm 1.19$ (hand--object interaction), $3.78 \pm 1.30$ (physical plausibility), and $3.57 \pm 1.31$ (training suitability). These results indicate that the synthetic videos exhibit high quality across all dimensions, particularly in instruction alignment and hand-object interaction, showing that WM-H synthetic videos can provide highly usable training supervision while maintaining good consistency with real manipulation videos in both visual and operational aspects.

Part C evaluates hand alignment after video editing. Participants rated pose consistency (C1) and contact preservation (C2) at $4.30 \pm 0.85$ and $4.25 \pm 0.84$, respectively, indicating that edited hands largely preserve the pre-edit pose and interactions.

\begin{table}[t]
  \centering
  \small
  \begin{tabular}{@{}llrrc@{}}
    \toprule
    Part & Dimension & $N$ & Result & $\Delta$ from real \\
    \midrule
    \multirow{4}{*}{A} 
      & AI judged as real & 355 & 37.7\% & -- \\
      & Judgment cue: visual smoothness / artifacts & 710 & 19.4\% & -- \\
      & Judgment cue: hand appearance \& motion & 710 & 34.9\% & -- \\
      & Judgment cue: physics, layout \& contact & 710 & 45.6\% & -- \\
    \midrule
    \multirow{5}{*}{B}
      & Object correctness & 1{,}420 & $3.97 \pm 1.22$ & 1.03 \\
      & Instruction alignment & 1{,}420 & $4.18 \pm 1.09$ & 0.82 \\
      & Hand-object interaction & 1{,}420 & $3.95 \pm 1.19$ & 1.05 \\
      & Physical plausibility & 1{,}420 & $3.78 \pm 1.30$ & 1.22 \\
      & Training suitability & 1{,}420 & $3.57 \pm 1.31$ & 1.43 \\
    \midrule
    \multirow{2}{*}{C}
      & Pose consistency & 355 & $4.30 \pm 0.85$ & -- \\
      & Contact preservation & 355 & $4.25 \pm 0.84$ & -- \\
    \bottomrule
  \end{tabular}
  \caption{User study results on hand-manipulation videos ($N=72$). Part~A: 10 videos (5 real + 5 AI); Part~B: 20 WM-H synthetic videos; Part~C: 5 before/after edit pairs. Part~A judgment cues sum to 100\%, Part~B/C rated on a 5-point Likert scale (1=worst, 5=best).}
  \label{tab:user-study}
\end{table}

\subsection{Failure Cases}

\begin{figure}[ht]
    \centering
    \includegraphics[width=1.0\linewidth]{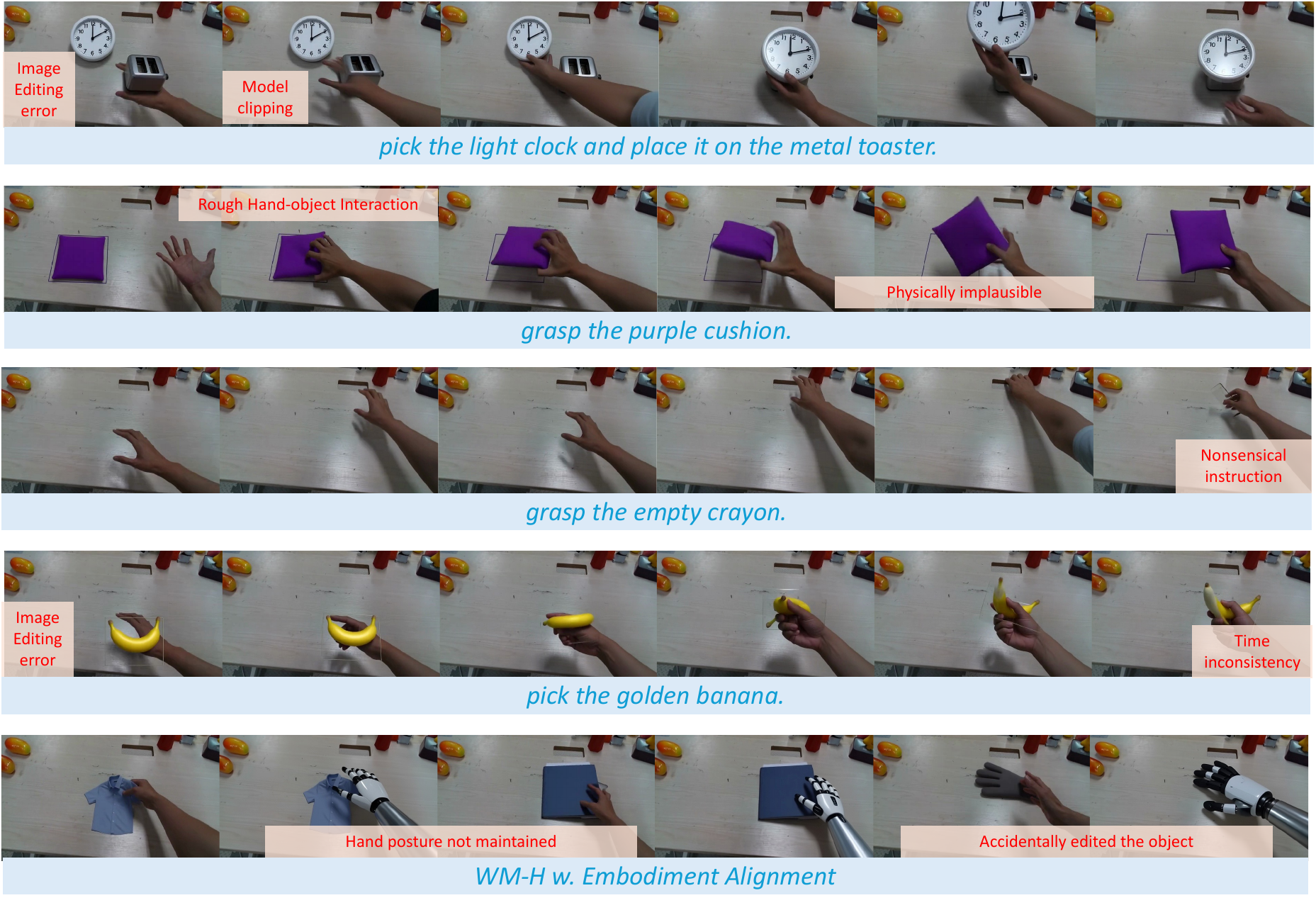}
    \caption{\small Qualitative visualization of representative WM-H failure cases. Each panel highlights typical issues such as image editing errors, physically implausible hand-object interactions, temporal inconsistencies, instruction misalignment, and imperfect robot-hand embodiment alignment.}
    \label{fig:wmh-failure-cases}
\end{figure}

Figure~\ref{fig:wmh-failure-cases} illustrates representative failure cases observed in WM-H-generated egocentric manipulation data, highlighting common issues that arise during video generation and human-to-robot hand alignment:

\begin{itemize}
    \item \textbf{Image Editing Errors:} Some frames exhibit incorrect object placement, visual artifacts, or cropping issues, which may mislead the policy during training.
    \item \textbf{Physically Implausible Interactions:} Certain actions violate real-world physics, such as hands passing through objects or unrealistic grasp poses, reducing the reliability of the supervision signal.
    \item \textbf{Inconsistent Pre- and Post-Interaction Visuals:} In a few cases, the visual state before and after the interaction is inconsistent, causing key interaction points to appear misaligned or resulting in incoherent action effects.
    \item \textbf{Invalid Instructions Leading to Unexecutable Actions:} Occasionally, assembled instructions lack practical meaning, making it impossible to execute the corresponding actions in the video, leading to operations that do not match the intended task.
    \item \textbf{Hand Editing Issues:} During robot-hand editing, the hand posture may not be preserved, or scene objects may be inadvertently modified, even though only the hand appearance was intended to be edited.
\end{itemize}

These failure modes indicate that, although WM-H provides scalable and diverse supervision data, careful attention is still required to ensure physical plausibility, visual fidelity, pre- and post-interaction consistency, instruction validity, and accurate embodiment alignment when using world-model-generated human-hand data for robot manipulation training.

\section{Policy Training Details}
\label{sec:APP-Policy}

\subsection{Architecture and Conditioning.}  
The policy uses a PaliGemma2-3B vision-language backbone to encode the current observation and language instruction. In addition to standard token embeddings, a 2D FoV token is projected via an MLP to the backbone hidden size and included in the input sequence. A separate, learnable cognition token is appended to the input; its final hidden state is extracted as the conditioning feature for the action decoder. The current hand state is represented in the camera frame with wrist translation, Euler angles, and 15-DoF MANO joint rotations for each hand (right-hand is the main focus). The action decoder is a diffusion-based DiT-B network, predicting 16 future steps per chunk with action dimension 102 corresponding to the MANO hand space.

\subsection{Human-Robot Action Alignment.}  
Robot demonstration states and actions are first converted from the robot's base frame to the camera frame. Wrist rotations are corrected to align with MANO conventions, and robot joint angles are retargeted to the human hand/MANO space. Although the policy internally pads state and action to a unified VITRA space (state dimension 212, action dimension 102), only the human-hand-related dimensions are active. Robot action and state values are normalized using statistics computed from human video data rather than robot-specific normalization. WM-H EA frames (edited WM-H for embodiment alignment) are used as additional supervision signals during training.

\subsection{Optimization and Inference}
During post-training, the vision encoder is frozen, while the remaining backbone and diffusion action decoder are updated. 
Training is conducted on 4 NVIDIA H200 GPUs with a per-GPU batch size of 64, giving a total batch size of 256. 
The learning rate is $1\times10^{-5}$ for both the backbone and action decoder, with weight decay 0.1, optimizer betas $(0.9, 0.95)$, gradient clipping 1.0, and a maximum of 40k training steps. 
The diffusion decoder uses 100 diffusion steps with a squared-cosine noise schedule. 
For each batch, we repeat diffusion training 8 times with independently sampled noise and timesteps. 
No image augmentation is applied. 
At inference, the policy is deployed on a single NVIDIA RTX 4090 GPU, using DDIM sampling with 10 steps and classifier-free guidance with scale 5.0.

\subsection{Data Mixture and Sampling Ratios}
Table~\ref{tab:data_mixture} summarizes the per-batch sampling ratios for different training setups. 
Here, $R$ denotes teleoperated robot data, $W$ denotes WM-H, and $W$-EA denotes embodiment-aligned WM-H. 
For HOI4D, we follow an episode-level annotation protocol: every 100 frames are grouped as one episode and assigned one language annotation, resulting in 5,511 annotated HOI4D episodes. 
The VITRA Real Version baseline is then trained with a balanced mixture between teleoperated robot data and these HOI4D episodes.

\begin{table}[h]
\centering

\label{tab:data_mixture}
\begin{tabular}{l c}
\toprule
Model / Setting & Dataset Sampling Ratio \\
\midrule
$\pi_{0.5}$ & $R = 1$ \\
VITRA & $R = 1$ \\
VITRA Real Version & $R:\mathrm{HOI4D} = 1:1$ \\
w/o scene alignment & $R:W\text{-EA}:W = 0.28:0.04:0.68$ \\
w/o embodiment alignment & $R:W = 0.4:1$ \\
WM-H 5k & $R:W\text{-EA}:W = 1:0.06:0.94$ \\
WM-H 25k & $R:W\text{-EA}:W = 0.28:0.04:0.68$ \\
\textit{PaliGemma pretrain}, Teleop & $R = 1$ \\
\textit{PaliGemma pretrain}, Teleop + WM-H & $R:W\text{-EA}:W = 0.28:0.04:0.68$ \\
\textit{Human pretrain}, Teleop & $R = 1$ \\
\textbf{Wh0 (50k)} & $R:W\text{-EA}:W = 0.28:0.04:0.68$ \\
\bottomrule
\end{tabular}

\caption{Training dataset composition for various models and ablations. $R$ denotes teleoperated robot data, $W$ denotes WM-H, and $W$-EA denotes embodiment-aligned WM-H.}
\end{table}

\section{Evaluation Details}
\label{sec:APP-EVAL}

\subsection{Real-World Experimental Setup}
\label{sec:app-real-world-setup}

\paragraph{Task suite.}
We evaluate all methods on a real Unitree G1 humanoid equipped with Inspire dexterous hands and an egocentric camera. The benchmark contains 18 real-world dexterous manipulation tasks:
(1) grasp the tripod,
(2) put the coke can into the black box,
(3) touch the robot gripper,
(4) put the glove into the orange container,
(5) grasp the paper cup,
(6) put the water bottle into the blue bowl,
(7) put the apple into the orange container,
(8) take the apple out of the orange container,
(9) grasp the towel,
(10) put the teapot into the orange container,
(11) wipe the table,
(12) put the white-green drink into the drawer,
(13) put the coke can into the drawer,
(14) grasp the remote controller,
(15) grasp the bread,
(16) put the tissue into the yellow basket,
(17) put the apple into the yellow basket, and
(18) grasp the tape measure.
All methods are evaluated with the same robot hardware, camera viewpoint, and low-level control interface. We report the mean success rate across all tasks along with the standard deviation to capture performance variability across the 18 tasks.

\paragraph{Stage-conditioned instruction protocol.}
For tasks involving multiple manipulation phases, we use stage-conditioned language instructions during evaluation. The instruction is matched to the current hand-object interaction state, such as reaching, grasping, or placing. When the robot completes one manipulation phase and enters the next, the evaluator provides the next instruction only; no manual control or action correction is applied. This protocol follows our data annotation scheme, where demonstrations are labeled according to distinct hand-object states. It allows the evaluation to focus on execution under the given visual observation and language condition, while keeping the protocol consistent across all methods.

\paragraph{Deployment-time action prior.}
Dexterous-hand control has a high-dimensional action space, and raw hand predictions can contain short-term noise during deployment. We therefore apply a simple grasping prior during the pre-contact phase: finger joints are constrained to move monotonically toward closure until a stable grasp is reached. This suppresses premature hand-opening motions near the object and improves real-robot stability. The same deployment rule is applied to all evaluated methods.

\subsection{Hand-Object Distance Evaluation}
\label{sec:app-ho-distance}

We use Hand-Object Distance as a proxy metric for language-conditioned object grounding. The evaluation set is constructed by using an LLM to generate unseen manipulation instructions and an image-editing model to synthesize unseen objects. We then estimate the 3D target object position from world-model trajectories. After manually filtering incorrect episodes, the final evaluation set contains around 5k episodes.
To obtain the target position, we identify key interaction frames from each world-model trajectory using local minima in hand velocity and changes in grasp/release state. We reconstruct hand motion at these frames and associate the recovered finger positions with the target object location.

At test time, we measure the distance between the policy-predicted wrist action and the estimated target object position. Lower distance indicates better instruction-conditioned object grounding. Since this metric does not evaluate finger-level dexterity, it should be viewed as a grounding metric rather than a complete task-success metric: low distance is usually necessary, but not sufficient, for successful manipulation.

\subsection{Robot Execution Results}

\begin{figure}[t!]
    \centering
    \includegraphics[width=1.0\linewidth]{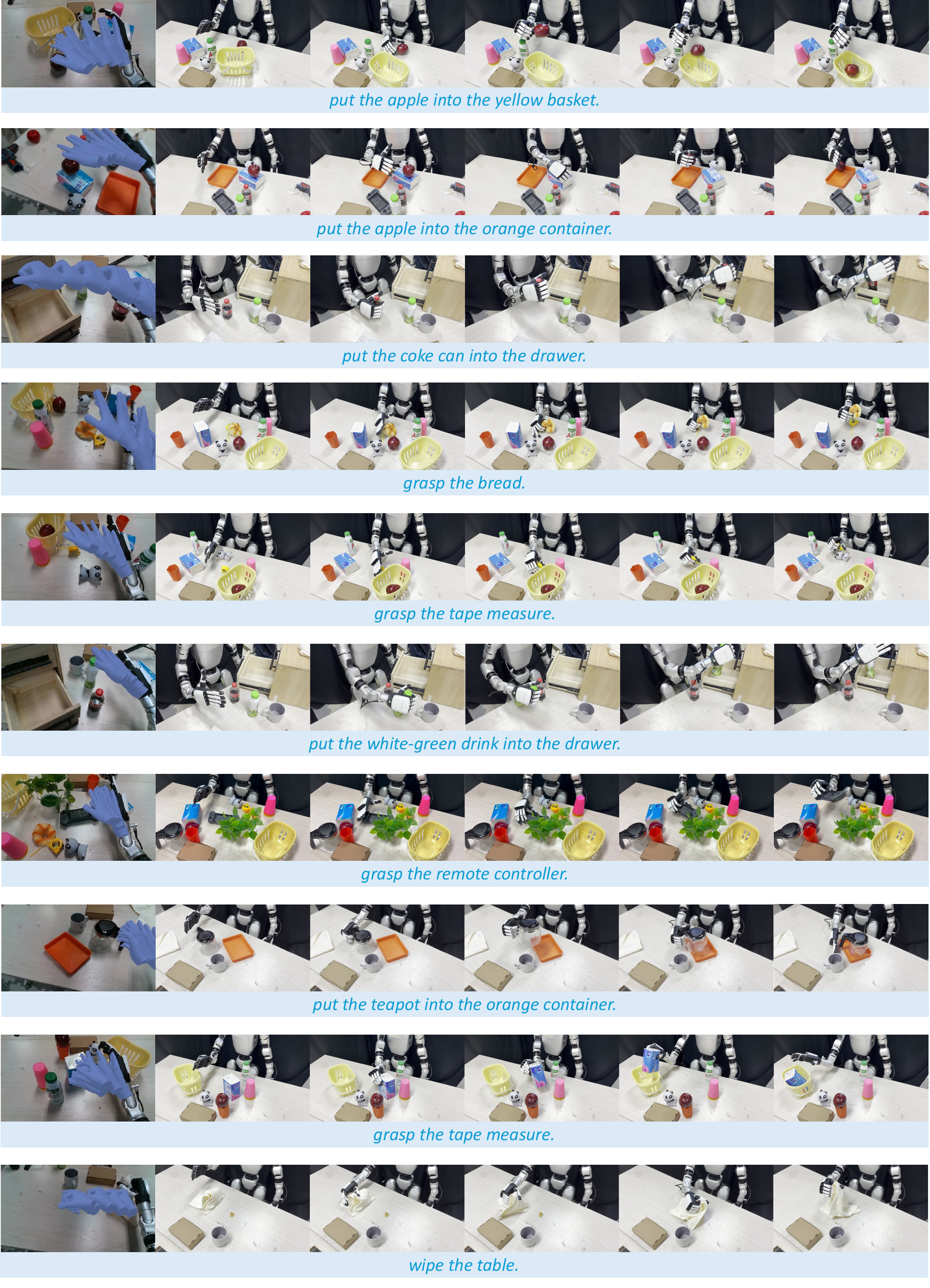}
    \caption{\small Robot execution rollouts of Wh0 across various dexterous manipulation tasks.}
    \label{fig:real-tasks-1}
\end{figure}

\begin{figure}[t!]
    \centering
    \includegraphics[width=1.0\linewidth]{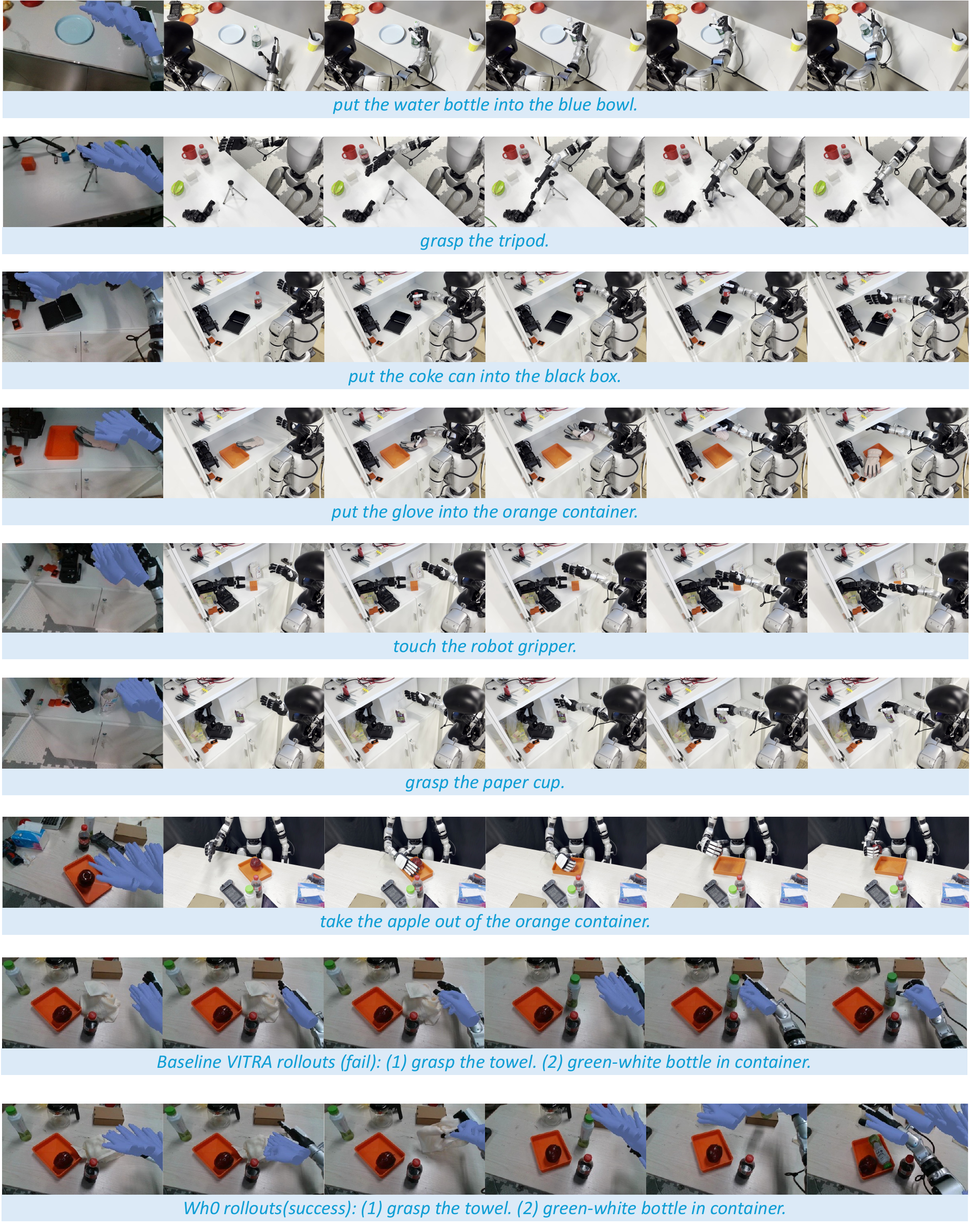}
    \caption{\small Additional Wh0 rollouts and comparison with baseline VITRA.}
    \label{fig:real-tasks-2}
\end{figure}

Figures~\ref{fig:real-tasks-1} and~\ref{fig:real-tasks-2} show qualitative rollouts of Wh0 on real-world manipulation tasks. The examples include object grasping, container placement, and table interaction, demonstrating that Wh0 can follow language instructions and execute diverse actions reliably. Compared with the baseline VITRA policy, Wh0 produces more consistent hand-object interactions and better task alignment.

\end{document}